\pdfoutput=1

\documentclass[11pt]{article}

\usepackage[final]{acl}

\usepackage{times}
\usepackage{latexsym}

\usepackage[T1]{fontenc}

\usepackage[utf8]{inputenc}

\usepackage{microtype}

\usepackage{inconsolata}

\usepackage{amsmath}
\usepackage{mathtools}
\usepackage{amssymb}
\usepackage{amsthm}
\usepackage{tabularx}
\usepackage{multirow}
\usepackage{multicol}
\usepackage{booktabs}
\usepackage{siunitx}
\usepackage{adjustbox}
\usepackage{{makecell}}
\setcellgapes{3pt}\makegapedcells

\usepackage{array,collcell}
\newcommand\AddLabel[1]{%
  \refstepcounter{equation}
  (\theequation)
  \label{#1}
}
\newcolumntype{M}{>{\hfil$\displaystyle}X<{$\hfil}}
\newcolumntype{L}{>{\collectcell\AddLabel}r<{\endcollectcell}}

\usepackage{graphicx}
\graphicspath{ {./images/} }
\usepackage[dvips]{epsfig}
\usepackage{svg}
\usepackage{relsize}
\usepackage{subcaption}
\usepackage{kotex}
\newcommand\Tau{\mathrm{T}}
\newcommand{\prob}[3][]{p_{#1}(#2\rightarrow #3)}
\newcommand{\meanstd}[2]{#1_{\pm{#2}}}
\theoremstyle{plain}

\theoremstyle{definition}
\newtheorem{definition}{Definition}
\usepackage{algorithm}
\usepackage[noend]{algpseudocode}
\algnewcommand\algorithmicforeach{\textbf{for each}}
\algdef{S}[FOR]{ForEach}[1]{\algorithmicforeach\ #1\ \algorithmicdo}
\usepackage{qtree}

\title{Structural Optimization Ambiguity and Simplicity Bias \\in Unsupervised Neural Grammar Induction}

\author{Jinwook Park \and Kangil Kim\thanks{Corresponding author.} \\
  AI Graduate School \\
  Gwangju Institute of Science and Technology\\
  \texttt{jinwookpark@gm.gist.ac.kr, kangil.kim.01@gmail.com}\\
}

\begin{document}
\maketitle

\begin{abstract}

Neural parameterization has significantly advanced unsupervised grammar induction. 
However, training these models with a traditional likelihood loss for all possible parses exacerbates two issues: 1) \textit{structural optimization ambiguity} that arbitrarily selects one among structurally ambiguous optimal grammars despite the specific preference of gold parses, and 2) \textit{structural simplicity bias} that leads a model to underutilize rules to compose parse trees. These challenges subject unsupervised neural grammar induction (UNGI) to inevitable prediction errors, high variance, and the necessity for extensive grammars to achieve accurate predictions. This paper tackles these issues, offering a comprehensive analysis of their origins. As a solution, we introduce \textit{sentence-wise parse-focusing} to reduce the parse pool per sentence for loss evaluation, using the structural bias from pre-trained parsers on the same dataset.
In unsupervised parsing benchmark tests, our method significantly improves performance while effectively reducing variance and bias toward overly simplistic parses. Our research promotes learning more compact, accurate, and consistent explicit grammars, facilitating better interpretability.

\end{abstract}

\section{Introduction}
Grammar induction (GI) has been intensively studied due to its critical role in understanding, controlling, and utilizing the structural information of internal symbols called constituents in linguistic analysis.
Unsupervised approaches to GI are continuously studied for their benefits of avoiding structured data as parses, which requires enormous human efforts and expertise to generate supervised data. 
In the literature of this field, neural parameterization for representing grammars~\cite{kim2019compound} has shown significant improvement over traditional probabilistic models~\cite{klein2002generative,klein2001natural,clark2001unsupervised}.
Furthermore, advanced models that incorporate conventional approaches, such as lexicalization~\cite{zhu2020return,yang2021neural} and symbol expansion~\cite{yang2021pcfgs,yang2022dynamic} have been proposed.

However, these Unsupervised Neural Grammar Inductions (UNGIs) lack analyses of the traditional challenges, particularly regarding the training loss. 
Some studies, such as \citet{kim2019compound}, have highlighted the difficulty of optimization presented by the complex loss landscape in UGIs~\cite{lari1990estimation,carroll1992two,klein2001natural}.
Subsequent efforts~\cite{yang2021pcfgs,yang2022dynamic} have alleviated these issues by only enhancing expressiveness through over-parameterization rather than the cause analysis. 

In this paper, we aim to reveal two unhandled problems in UNGIs: 1) \textit{structural optimization ambiguity}, which increases the number of optima of the same quality that prefer different parse structures, leading to prediction errors by random selection for branching and high variance, and 2) \textit{structural simplicity bias}, which induces the grammar to use only a few rules in parse trees and resulting in simplistic parse trees even with a large capacity. 
Both problems are caused by training a neural grammar with sentence probability.

To address these issues, we propose a simple but effective method to inject focusing-bias that accounts for only a few selected parses for training loss in the inside algorithm.
This approach reduces the ambiguity by selecting more distinct parses across sentences and restricts the simplicity bias by excluding of unnecessary simplistic parse trees by sentences.
We also introduce a method of generating structural bias by parse selection from single or multiple parsers pre-trained on the same training data. 

Our method outperforms state-of-the-art implicit and explicit grammar induction models in English parsing tasks on the Penn Treebank (PTB)~\cite{marcus-etal-1993-building}, as well as in ten other languages, including the Chinese Penn Treebank (CTB)~\cite{xue2005penn} and the SPMRL datasets~\cite{seddah-etal-2014-introducing}.
In-depth analysis shows a reduction in high variance and overly simplistic parses. We provide the investigation results of various structural biases, recommending heterogeneous multi-parsers. 

Our contributions are summarized as follows:
\begin{itemize}
    \item We raise and clarify \textit{structural optimization ambiguity} and \textit{structural simplicity bias}, which lead to high variance and overly simplistic parse problems in UNGIs.
    \item We propose \textit{sentence-wise parse-focusing} that employs biases from pre-trained parsers using the same training data. This approach enables the stable learning of more compact and accurate explicit grammars.
    \item Through in-depth empirical analysis, we demonstrate the effectiveness of our approach in reducing the identified causes, investigating various focusing-biases, and achieving significant performance improvements compared to state-of-the-art UNGIs. 
\end{itemize}

\section{Background}

\subsection{Notations of Probabilistic Context-Free Grammar}
\label{sec:pcfgs}

In the following analysis, we use the notations below for a PCFG $G=(S, N, P, \Sigma, R, \Pi)$: the root symbol $S$, finite sets of nonterminals $N$, preterminals $P$, terminal words $\Sigma$, production rules $R$, and production rule probabilities $\Pi$. $R$ consists of three types of rules:
\begin{equation*}
    \begin{array}{ll}
        S \rightarrow A
        &\text{where } A\in N \\
        A \rightarrow\alpha\;\beta
        &\text{where }\alpha,\beta\in (N\cup P) \\
        B \rightarrow \omega
        & \text{where }B\in P\text{ and } \omega\in\Sigma
    \end{array}
\end{equation*}

Note that the left-hand side (parent) of the rules is distinguished between nonterminal and preterminal categories.

\subsection{Negative Log Likelihood Loss}
\label{sec:nll}

The loss of grammar induction is well-known to increase the probability mass of only observed parse trees~\cite{chi1998estimation} and to maximize rule probability for representing correct parses of sentences~\cite{corazza2006cross}.
The loss of UGI is the same as SGI in that they both maximize sentence probability for a given sentence; however, it is different in that UGI uses all possible parse trees to calculate sentence probability.

The likelihood loss for a given sentence is:
\begin{equation*}
    \begin{split}
        L(s) &
        = -\log \sum_{\tau\in\Tau(s)}\exp{\sum_{r\in R(\tau)}{f(r|\tau)\log p(r|G)}}
    \end{split}
    \label{eq:nll}
\end{equation*}

where the exponential term is the probability of a parse tree $\tau$, and $f(r|\tau)$ denotes the frequency of rules used in the tree.
In the case of SGI, $\Tau(s)$ only contains the gold parse tree for $s$. However, for UGI, $\Tau(s)$ contains all parse trees that can generate $s$.

In the equation~\ref{eq:nll}, the loss reaches its minimum when the exponential term within the summation, which is always negative or equal to zero, approaches zero. This can be rewritten as $\sum f(A\rightarrow\alpha)\log P(A|\alpha) \simeq 0$.

\section{Limits on Unsupervised Learning of Neural Grammar Induction}

As explained earlier, UNGI Loss has the advantage of inducing grammar without the information of gold parse trees.
However, UNGIs still inherit the limitations of probabilistic UGIs.
In this section, we reveal two issues arising from these limitations in UNGIs:\textit{structural optimization ambiguity} and \textit{structural simplicity bias}, and discuss derived negative effects and their causes.
For precise analysis, we focus on inducing explicit neural grammars based on PCFGs, mainly Neural PCFGs (N-PCFGs). 

\subsection{Structural Optimization Ambiguity (SOA)}

\textit{Structural optimization ambiguity} causes a huge variance of performance by arbitrary branching bias that depends on random seeds.
In this section, we clarify the definition of SOA, and explain the arbitrary convergence caused by SOA.
After that, we mathematically prove the existence of SOA under the single pre-terminal condition and extend this to the multiple pre-terminal condition using empirical evidences.

\paragraph{Definition of Structure Optimization Ambiguity by UGI Loss}
SOA is the ambiguity that disrupts the distinction between two optima that predict different parse trees for the same sentence.
For clear analysis, we define the ambiguity as follows.

\begin{definition} \label{def:ambiguity}
Given $G_1=(S,N,P,\Sigma,R,\Pi_1)$ and $G_2=(S,N,P,\Sigma,R,\Pi_2)$ that assign different rule probabilities from the same $G$, the grammar $G$ is \textit{structural-optimization-ambiguous} if and only if $p(s|G_1) = p(s|G_2)$ for all sentences $s\in S$ in the training data, and there exists a sentence $s$ whose best parses respectively derived through $G_1$ and $G_2$ are different. 
\end{definition}

As mentioned in Section~\ref{sec:nll}, the sentence probability is the loss function of GI.
This means that if two grammars have the same sentence probability, they are not distinguished in the optimization landscape.
In other words, they are ambiguous in the perspective of optimization.

\paragraph{Arbitrary Convergence by Structural Optimization Ambiguity}

\begin{figure}
    \centering
    \includegraphics[
        width=6cm
    ]{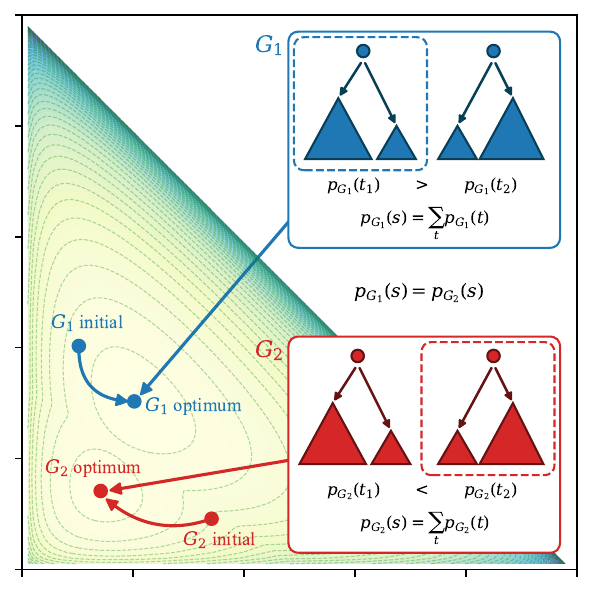}
    \caption{The optimization of two different grammars, $G_1$ and $G_2$, in the optimization landscape for a given sentence. Each grammar has different preferences for parse trees based on parse tree probability; however, they have the same sentence probability. The two initial points converge to the different optima close to their own.}
    \label{fig:structural-ambiguity}
\end{figure}

The discrepancy between structure and loss function leads to that each grammar randomly initialized favors different optimal points.
In Figure~\ref{fig:structural-ambiguity}, highlighted two optimal points are equally probable yet structurally divergent.
At the blue optimum, the parse tree $t_1$ has a higher probability than the parse tree $t_2$, whereas, at the red optimum, the situation is reversed, with $t_2$ having a higher probability than $t_1$.
Consequently, it becomes unclear which optimum will be chosen based solely on sentence probabilities from an optimization perspective.
Moreover, it is intuitively understood that the reachable optimum may vary with the randomness, such as initial states.

\paragraph{Proof in Single Pre-terminal Condition}

We prove that SOA always exists under the single pre-terminal condition by using equivalence of sentence probability between two grammars that have different probability distributions.


In the first step, we consider constraints on the grammars: 1) both grammars share the same configuration $(S,N,P,\Sigma,R)$, and 2) the rule probabilities of both grammars are identical except for a single type of flipped rule pair ($N_i\rightarrow N_j T$ and $N_i\rightarrow T N_j$).
Under these constraints, suppose that the two grammars yield the same sentence probability for a given sentence; we can derive the following equation in general condition from the formula of the inside algorithm:

\begin{equation}
\begin{aligned}
&\prob[G]{N_i}{N_jT}+\prob[G]{N_i}{TN_j}\alpha\\
&=\prob[G']{N_i}{N_jT}+\prob[G']{N_i}{TN_j}\alpha,\\
\end{aligned}
\label{eq:proof_soa_1}
\end{equation}
Where,
\begin{equation*}
\alpha=\frac{\prob[G_1]{T}{w_1}I_{G_1,j}(2,r)}{I_{G_1,j}(1,r-1)\prob[G_1]{T}{w_r}}
\end{equation*}

Then, we also consider that each grammar includes only one pre-terminal\footnote{The single pre-terminal condition is also established when 1) rule probabilities are uniformly initialized or 2) all tokens in sentences cannot be distinguished (e.g. all tokens are \textit{<unk>}).}.
Under the aforementioned constraints, $\alpha=1$ is always satisfied; therefore, the above equation~\ref{eq:proof_soa_1} can be simplified as follows:

\begin{equation}
\begin{aligned}
&\prob[G]{N_i}{N_jT}+\prob[G]{N_i}{TN_j}\\
&=\prob[G']{N_i}{N_jT}+\prob[G']{N_i}{TN_j}
\end{aligned}
\label{eq:proof_soa_2}
\end{equation}

Under the identical probability distribution except for the flipped rule pair, the sum of the probabilities for the flipped rule pair is always equal in a proper grammar \cite{nederhof2006estimation}.
Thus, equation~\ref{eq:proof_soa_2} is always satisfied, and the two grammars always yield the same probability for the same sentence.
Furthermore, due to the strict inequality of the flipped rule pair, the two grammars can derive different parse trees.
This signifies that the two grammars are structural-optimization-ambiguous.
Appendix~\ref{appendix:ambiguity-proof} provides a more detailed proof of this.



\paragraph{Empirical Evidence in General Condition: Low Correlation of Loss to S-F1}

Because proving the existence is significantly complex under general condition using multiple pre-terminals, we prove it indirectly by presenting empirical evidences.
The first evidence is a low correlation between sentence-level F1 (S-F1) scores\footnote{The S-F1 score is one of the common metrics to evaluate parsing quality. It evaluates the similarity between predicted parses and gold parses. In this paper, we only use unlabeled S-F1 scores that do not consider the symbol labels the same as N-PCFG.} and negative log likelihood loss, which implies that the optimization of grammars is inconsistent, and can lead to generating structures different from those desired by gold parses.

\begin{figure}
    \centering
    \begin{subfigure}[b]{0.48\columnwidth}
        \includegraphics[
            width=\columnwidth
        ]{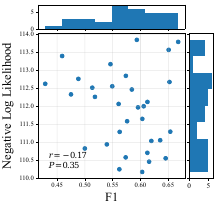}
        \caption{Correlation of S-F1 score and likelihood}
        \label{fig:f1-ll-corr-new}
    \end{subfigure}
    \hfill
    \begin{subfigure}[b]{0.48\columnwidth}
        \includegraphics[
            width=\columnwidth
        ]{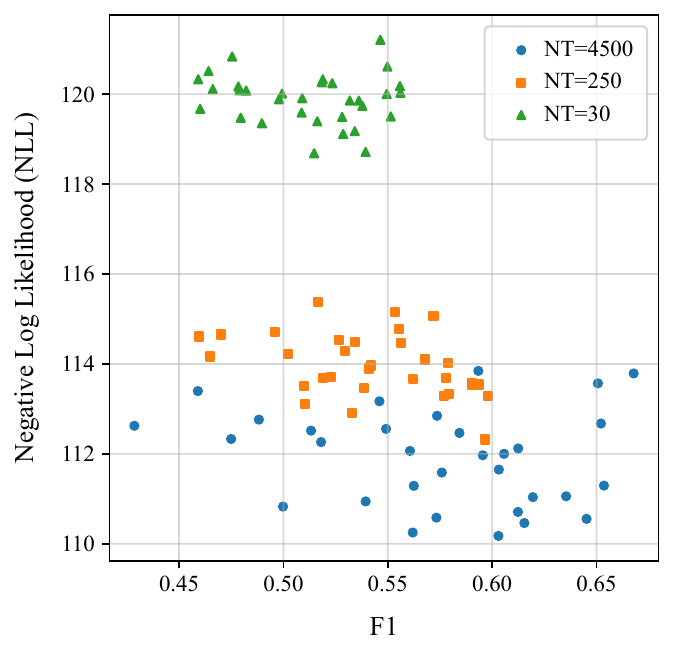}
        \caption{Amplification for the variance of S-F1 by symbol size}
        \label{fig:corr-for-sym-new}
    \end{subfigure}
    \label{fig:corr-set}
    \caption{(a) Low correlation between S-F1 score and UGI likelihood for PTB test set. We evaluate trained FGG-TNPCFGs with 4500 nonterminal symbols and 9000 preterminal symbols using 32 random seeds over 10 epochs. (b) Amplification of variance by symbol size increasing. Circles are the same as in (a) and others are trained with the same settings as in (a) except for the number of symbols.}
\end{figure}

In Figure~\ref{fig:f1-ll-corr-new}, the negative log likelihood and S-F1 scores are shown to be almost uncorrelated, as indicated by a low P-value and Pearson coefficient. 
This low correlation means that the distinction between the grammars that induce different structures is hard to determine using negative log likelihood, which causes the large variance of S-F1 scores due to randomness.
Through this analysis, we demonstrate the existence of SOA under general conditions.

\paragraph{Empirical Evidence in General Condition: Variance Amplification by Symbol Size}
Another piece of evidence is the variance amplification of S-F1 scores with the increase in the number of symbols, as shown in Figure~\ref{fig:corr-for-sym-new}. 
It appears that more nonterminals create more local optima of similar quality that derive different parses, leading to a greater variety of parses compared to a single gold parse.
This demonstrates the variance amplification of SOA as the number of symbols increases under general conditions.

\subsection{Structural Simplicity Bias (SSB)}
\label{sec:simplicity}
The second problem, \textit{structural simplicity bias}, causes a decrease in expressive power for parse trees by leading the grammar to have low structural diversity.
In this section, we discuss what SSB is, the problems it causes, and why it occurs.
Lastly, we present empirical evidence for how SSB is revealed during training.

\paragraph{What is SSB?}


SSB is the bias that induces a grammar to utilize as few rules as possible to compose parse trees.
This bias is a natural and intended effect of the objective to maximize sentence probability.
However, it leads to learning undesirable structures that are different from gold parses in UNGI.

\paragraph{Specific Conditions of UGI}
UGI is affected by different biases, unlike SGI, due to two properties: 1) the loss maximizes the probabilities of all possible trees for a given sentence. 2) all rules can be used in any position that has the same type\footnote{The type of rules means the four types, $N\rightarrow N N | N P | P N | P P$, distinguished by the type of symbols: nonterminal, preterminal.}.
Therefore, UGI prefers the structure that maximizes tree probability rather than optimizing the given parses as SGI does.

\paragraph{Rule Simplification for Expressing a Parse}
If all rules in the tree are the same, it is intuitively easy to maximize tree probability because the model only needs to optimize one rule.
Therefore, maximization using tree probability prefers to train utilizing a small number of rules. This preference leads to a grammar that includes simple structures in parses.
This means that the grammar has low expressive power for parse trees.
Specifically, only rules of same type can share their position in the tree due to symbol types; therefore, parses prefer structures that utilize structural repetition. For these reasons, it prefers left- or right-binarized structures with many repetitions.

\paragraph{Space Inefficiency and Performance Degradation by the Potential Expressive Limit}

The grammar affected by SSB utilizes only a small part of its potential expressive power for parse trees, which causes two problems: 1) space inefficiency, as expressive power grows slowly compared to an increase in grammar size, and 
2) low expressive power compared to gold parses, leading to decreased performance.
The low rule diversity of grammar compared to the necessity of gold parses cannot represent the necessary tree structures for a given sentence.
Induced parses that are not sufficiently represented cause low performance.

\paragraph{Empirical Evidence: Average Diversity in Each Parse Tree}

\begin{figure}
    \centering
    \begin{subfigure}[b]{0.48\columnwidth}
        \includegraphics[
            width=\columnwidth,
        ]{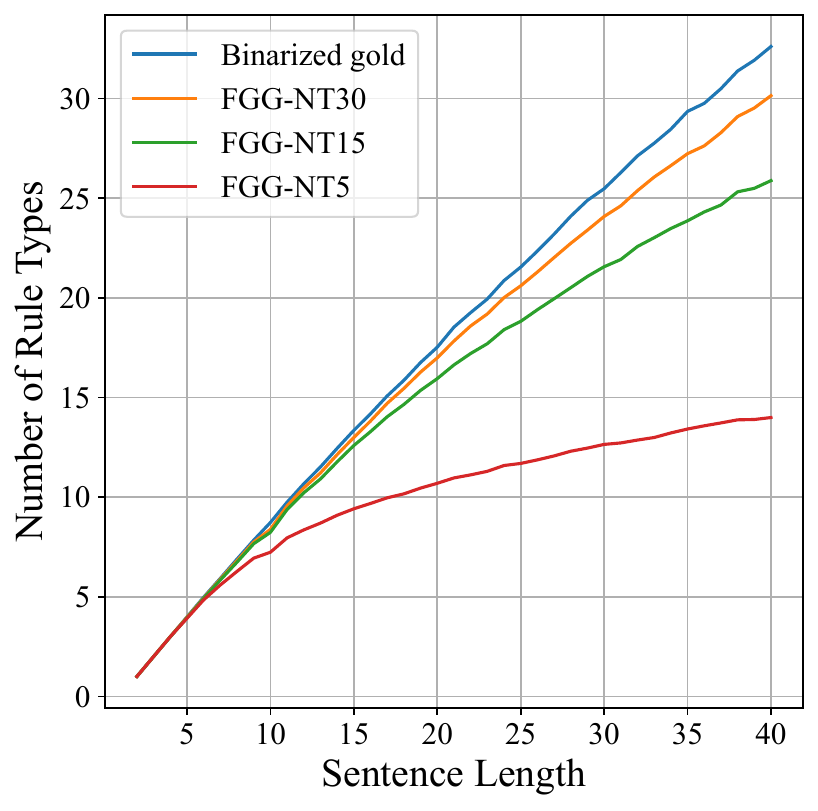}
        \caption{Comparison between the gold parse trees and FGG-TNPCFGs.}
        \label{fig:util_fgg}
    \end{subfigure}
    \hfill
    \begin{subfigure}[b]{0.48\columnwidth}
        \includegraphics[
            width=\columnwidth,
        ]{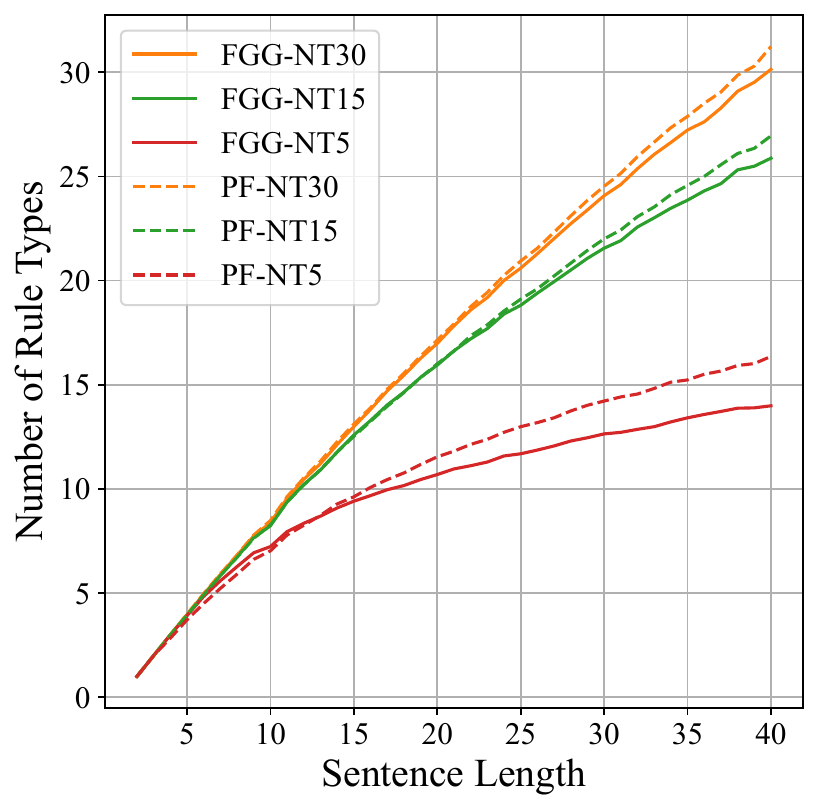}
        \caption{Comparison between FGG-TNPCFGs and parse-focused N-PCFGs.}
        \label{fig:util_comp}
    \end{subfigure}
    \caption{The average number of unique rules in each parse for each sentence length. We evaluate the average for models with 5 / 15 / 30 nonterminals trained with four different random seeds. These evaluations use the WSJ train set. (a) The gold uses non-binarized gold parses. (b) Parse-focused N-PCFGs use models trained by Structformer, NBL-PCFG, and FGG-TNPCFG.}
\end{figure}

We demonstrate that existing models induce simplified grammars by analyzing the predicted parses.
We provide evidence that SSB leads a model to learn simplistic diversity compared to gold parses, as shown in Figure~\ref{fig:util_fgg}, which simplifies the induced tree structures and results in similar structures among trees of the same length. In other words, this can damage diversity.
This decrease in diversity appears to lead to a decrease in performance.

In Figure~\ref{fig:util_comp}, the grammars of the same size with parse-focusing have greater rule diversity than FGG-TNPCFGs.
This shows that 1) parse-focusing relaxes the SSB, and 2) the decrease in rule diversity is not caused by the upper bound for grammar capacity.
We verify the loss of rule diversity in other languages in Appendix~\ref{appendix:rule_util_multilingual} and come to the same conclusion.
Additionally, we provide the qualitative analysis of overly simplistic parses in Appendix~\ref{appendix:qa_simplicity}.

\section{Method}
\subsection{Preliminaries}
\label{sec:preliminaries}

\paragraph{Motivation}
SOA and SSB have not yet been reported as critical problems, even though they can also exist in conventional probabilistic GIs~\cite{klein2001natural}.
The reason is that conventional methods can avoid equal frequency assignment to all parses, which is the basic condition that causes these problems. 
For example, the EM algorithm in UGIs reassigns different rule frequencies at every iteration~\cite{petrov2006learning}, and SGIs use only gold parse trees while implicitly assigning zero probability to all incorrect parses. 
Motivated by this, we propose \textit{sentence-wise parse-focusing} to learn from a few parses for each sentence and \textit{structural bias generation} to select the parses from unsupervised multi-parsers pre-trained on the same training data.

\paragraph{Overview of Parse-Focusing}
\begin{figure}
    \centering
    \includegraphics[
        width=\columnwidth
    ]{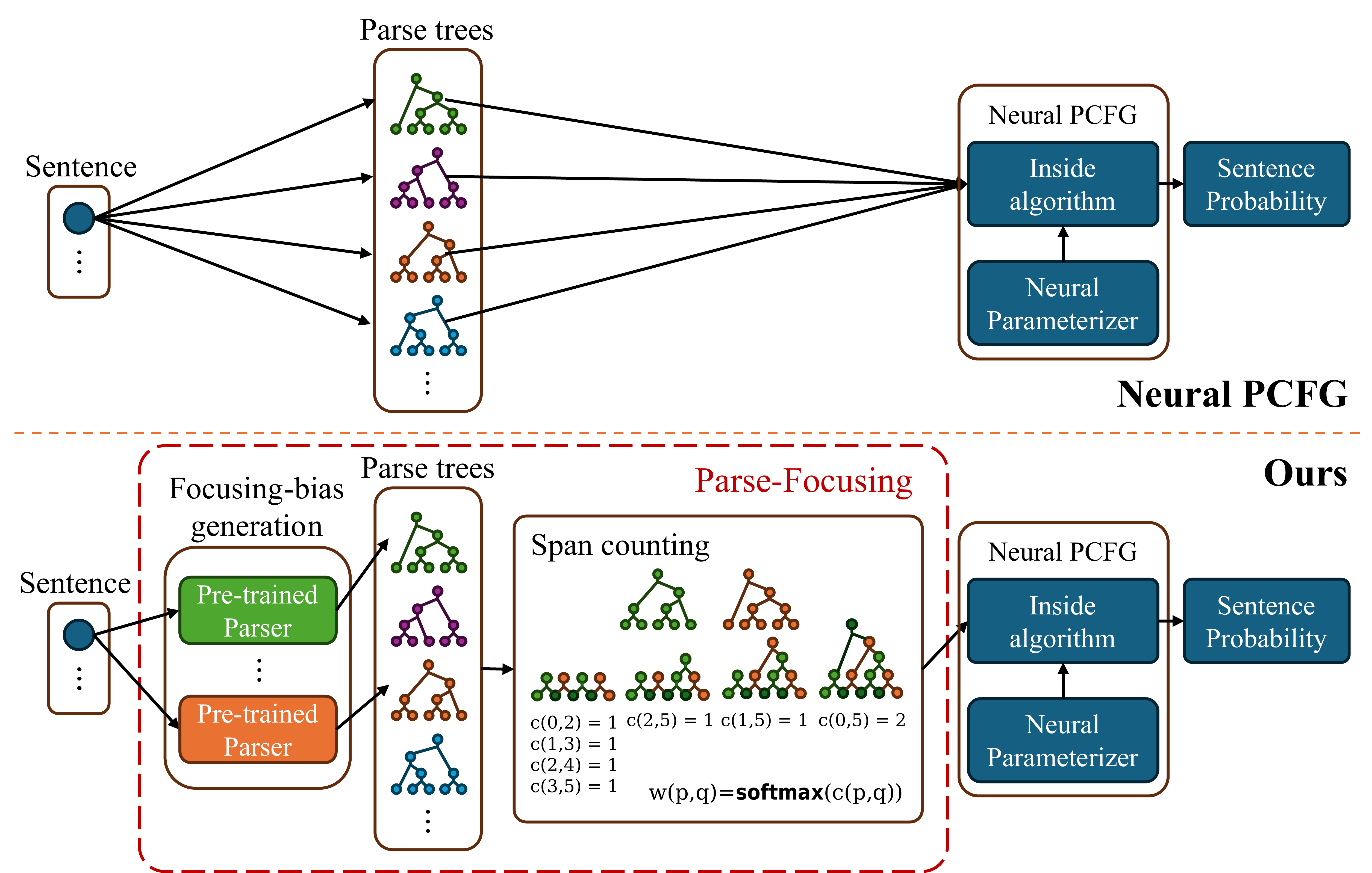}
    \caption{Overview of N-PCFG (above) and Ours (below). The red box shows the parse-focusing method.}
    \label{fig:overview}
\end{figure}

Figure~\ref{fig:overview} presents an example application of sentence-wise parse-focusing to N-PCFGs. 
Compared to N-PCFG, the distinguishing feature is selecting a few parses from unsupervised parsers pre-trained on the same training data and deriving the training loss from these parses.
Details of the algorithm are provided in Appendix~\ref{appendix:alg-method}.

\subsection{Sentence-wise Parse-Focusing}
\paragraph{Learning from Focused Parses}
To learn grammar from a few selected parses, we simply modify the inside algorithm to count constituent frequency only for observed parses.
However, this strict bias injection completely excludes unselected parses, resulting in losing potential structures. 
To alleviate this limitation, we use soft weights derived from softmax over all constituent counts, which maintains the focus on the selected parses and also leaves small weights on potential structures. 
The process of applying the parse-focusing is detailed in Algorithm~\ref{alg:overview}.

Given a finite set of sentences $s\in\mathcal{S}$, we represent a finite set of selected parse trees for each sentence $s$ as $\tau_{s}\in\mathcal{T}_{s}$.
We denote the frequency for span $(p,q)$ of word $p$ to $q$ for parse tree $\tau_{s}$ as $f((p,q)|\tau_{s})$.
Then, the soft weights are 

\begin{equation*}
    w(p,q|s)=\frac{\exp\sum_{\tau_{s}\in\mathcal{T}_{s}}{f((p,q)|\tau_{s})}}{\sum_{p,q}\exp \sum_{\tau_{s}\in\mathcal{T}_{s}}f((p,q)|\tau_{s})}
\end{equation*}

\paragraph{Why Does It Alleviate High Variance and Overly Simplistic Parse Issues?}
As mentioned in Section~\ref{sec:preliminaries}, the base condition for causing SOA and SSB is that all parse trees are considered equally.

This equal consideration establishes conflicting pairs of flipped rules with equal frequency, which causes SOA.
Sentence-wise parse-focusing avoids this equal frequency by using soft weights for spans.
Additionally, sentence-wise parse-focusing has no specific dynamics to consider them equally.
These reasons significantly decrease the possibility of establishing Equation~\ref{eq:proof_soa_1} under general condition, which avoids the problems caused by SOA in training.

This consideration also makes a model maximize advantageous simplistic parse trees that are not related to gold parse trees, which causes SSB.
These simplistic parse tree structures are shared across sentences.
Sentence-wise parse-focusing avoids the maximization of the simplistic parse trees by assigning selected parse trees for given sentences.
The selected parse trees may not involve the simplistic parse trees and are also different across sentences.

\subsection{Focusing-Bias Generation}
\paragraph{Bias Generation from Pre-Trained Unsupervised Parsers}
The performance of grammar relies on the quality of selected parses because focused parses control the loss for given sentences.
Therefore, we can expect to further improve performance when we use high-quality parses.
However, in unsupervised learning, finding parses close to gold parse trees without supervision-related information is a challenging issue. 
As an alternative solution, we use additional parsers pre-trained on the same data with unsupervised learning. Note that our method allows any type of parser as long as it provides parse span information.

\paragraph{Data-Specific Bias Induction via Heterogeneous Multi-Parsers}

A potential approach to generate good parses is to use parsers induced by different models, which we call heterogeneous parsers. 
According to \citet{ishii-miyao-2023-tree}, the branching bias consists of model-specific inherent bias and data-specific inductive bias.
They also show that different parsers have different model-specific inherent biases.
Therefore, if we use heterogeneous parsers when counting constituents in parse-focusing, the opposing inherent biases between parsers are offset, while the same structures by data-specific bias are enhanced.

In Table~\ref{tab:heteroparsers-iou}, we compare the structural differences between parsers within pairs of heterogeneous parsers and within pairs of homogeneous parsers.
Homogeneous parsers are trained using the same model but with different random seeds.
To directly quantify structural heterogeneity, we use IoU scores on parse spans defined as:

\begin{equation*}
    \frac{\bigcap_{G_p\in\mathcal{G}_{p}}{\Tau_{G_p}}}{\bigcup_{G_p\in\mathcal{G}_{p}}{\Tau_{G_p}}}
\end{equation*}

The IoU scores of heterogeneous parsers are consistently lower than those of homogeneous parsers, which implies that these parsers have more different branching biases.
Therefore, when using heterogeneous parsers, the different model-specific inherent biases are offset, making the shared part more likely to indicate data-specific bias.

\begin{table}[t!]
    \tiny
    \centering
    \begin{tabular}{lc|lc}
    \hline
    \multicolumn{2}{c}{Homogeneous 
        } & \multicolumn{2}{c}{Heterogeneous} \\
    \hline
        Parser Pair & IoU & Parser Pair & IoU \\
    \hline
        (SF, SF) & $\meanstd{56.3}{7.0}$ & (SF, NBL) & $\meanstd{32.8}{9.8}$ \\
        (FGG, FGG) &  $\meanstd{47.0}{7.3}$ & (SF, FGG) &        $\meanstd{38.9}{5.6}$ \\ 
        (NBL, NBL)  & $\meanstd{38.5}{19.7}$ &
        (NBL, FGG) &  $\meanstd{37.9}{12.3}$ \\
    \hline
    \end{tabular}
    \caption{IoU scores of homogeneous and heterogeneous parser combinations. SF means Structformer, FGG means FGG-TNPCFGs and NBL means NBL-PCFGs.}
    \label{tab:heteroparsers-iou}
\end{table}

\section{Experiments}

\subsection{Settings}

\paragraph{Datasets and Hyperparameters}

We use the Penn Treebank (PTB)~\cite{marcus1994penn}, an English constituency parsing task dataset, to evaluate parsing performance. Additionally, we use the CTB~\cite{xue2005penn} and SPMRL dataset~\cite{seddah-etal-2014-introducing} for multilingual evaluation.

We use the same hyperparameters as \citet{yang2022dynamic}; for the analysis of the number of symbols, we follow the ratio of 1:2 between nonterminals and preterminals. The details for the dataset and training are in Appendix~\ref{appendix:exp_details}

\paragraph{Inference}
We primarily evaluate grammars with MBR decoding~\cite{yang2021pcfgs}. However, we use CYK decoding when it is necessary to distinguish each rule in the grammar.

\paragraph{Base Model}
We use the model from \citet{yang2022dynamic}, which we refer to as Factor Graph Grammar-based TN-PCFGs (FGG-TNPCFGs), as our base model. Our model is implemented based on the open-sourced code from~\citet{yang2022dynamic}\footnote{https://github.com/sustcsonglin/TN-PCFG}. There are two reasons for choosing this model:
1) The strict assumption of PCFGs is preserved.
In the case of Compound PCFGs (C-PCFGs) and Lexicalized PCFGs~\cite{zhu2020return,yang2021neural}, because they are dependent on words or sentences, which makes the induced grammar complex to understand.
We avoid this to verify whether better performance can be achieved without relaxed assumptions.
2) FGG-TNPCFGs rapidly train large-scale grammars through efficient use of memory and computation. This is favorable for obtaining diverse results in different conditions. Additionally, we note that while FGG-TNPCFGs provide efficient computation, the fundamental mathematical principle for grammar remains the same.

\paragraph{Selected Parsers for Focusing-Bias}
We adopt three parsers that use different algorithms;
1) Structformer~\cite{shen2021structformer} is a transformer-based and implicit grammar model.
2) NBL-PCFG~\cite{yang2021neural} is a lexicalized explicit grammar model.
3) FGG-TNPCFG~\cite{yang2022dynamic} is the largest and state-of-the-art grammar model.
Note that all of these parsers use the same hyperparameter settings as in their original papers, and they are trained only with PTB. The primary model trained by parse-focusing consists of these three parsers, which shows the best performance.
Additionally, we use DIORA~\cite{drozdov2019unsupervised}, which is a self-supervised model utilizing a recursive autoencoder architecture.
However, it is not used as the main parser because it is trained with the additional dataset MultiNLI~\cite{williams-etal-2018-broad} besides PTB.

\subsection{Performance}

\paragraph{Performance in English}

We evaluated the performance of state-of-the-art UNGI methods on English PTB data, as shown in Table~\ref{tab:final-performance}. 
Our approach significantly outperforms the state-of-the-art model (FGG-TNPCFGs) with only 30 nonterminals and 60 preterminals.
When using 4500 nonterminals, as with FGG-TNPCFGs, the performance further increases significantly. 
Note the difference between reproduced models and the variance among models. Instead of running only four runs, we performed 32 runs, resulting in lower performance and higher variance, especially in FGG-TNPCFG and NBL-PCFG.
The significant performance difference between DIORA and DIORA$^\dag$ is due to differing evaluation criteria. While DIORA utilizes a test set of binarized trees by Stanford CoreNLP, DIORA$^\dag$ employs non-binarized trees. This approach is intended to assess performance accurately within the evaluation environment of N-PCFG.

\begin{table}
    \small
    \centering
    \begin{tabular}{clcc}
        \hline
        \multirowcell{2}{grammar\\type}&\multicolumn{1}{c}{\multirow{2}{*}{model}} & \multicolumn{2}{c}{S-F1} \\
        \cline{3-4}
        && mean & max \\
        \hline
        \multirowcell{4}{\\implicit\\grammar}& Structformer & 54.0 & - \\
        &Structformer$^{\dag}$ & $\meanstd{52.3}{2.3}$ & 54.2\\
        &DIORA & 55.7 & 56.8 \\
        &DIORA$^{\dag}$ & $\meanstd{43.6}{0.9}$ & 44.8 \\
        \hline
        \multirowcell{12}{\\\\\\\\explicit\\grammar}&N-PCFG & $\meanstd{50.9}{2.3}$ & 52.3 \\
        &N-PCFG w/ MBR & $\meanstd{52.3}{2.3}$ & 55.8 \\
        &C-PCFG & $\meanstd{55.4}{2.2}$ & 59.0 \\
        &C-PCFG w/ MBR & $\meanstd{56.3}{2.1}$ & 60.1 \\
        &NL-PCFG & 55.3 & - \\
        &NBL-PCFG & 60.4 & - \\
        &NBL-PCFG$^{\dag}$ & $\meanstd{53.3}{11.5}$ & 62.3 \\
        &TN-PCFG & $\meanstd{57.7}{4.2}$ & 55.6\\
        &FGG-TNPCFG & 64.1 & - \\
        &FGG-TNPCFG$^{\dag}$ & $\meanstd{57.4}{6.0}$ & 66.8 \\
        \cline{2-4}
        & Ours (NT=30) & $\meanstd{67.4}{0.9}$ & 68.4 \\
        & Ours (NT=4500) & $\meanstd{69.6}{0.6}$ & 70.3 \\
        \hline
        
    \end{tabular}
    \caption{Performance of UNGIs on English trained on Penn TreeBank (\dag: reproduced result). All statistics are from 32 runs with different random seeds. Our model is trained with randomly and separately generated focusing-biases from Structformer$^\dag$, NBL-PCFG$^\dag$, and FGG-TNPCFG$^\dag$.
     }
    \label{tab:final-performance}
\end{table}

\paragraph{Performance in Multilingual}

\begin{table*}
    \centering
    \begin{adjustbox}{max width=\textwidth}
    \begin{tabular}{lccccccccccc}
    \hline
        Model & Basque & Chinese & English & French & German & Hebrew & Hungarian & Korean & Polish & Swedish & Mean rank\\
    \hline
        N-PCFG* &
        $\meanstd{35.1}{2.0}$ & $\meanstd{26.3}{2.5}$ & $\meanstd{52.3}{2.3}$ & $\meanstd{45.0}{2.0}$ & $\meanstd{42.3}{1.6}$ & $\meanstd{45.7}{2.2}$ & $\meanstd{43.5}{1.2}$ & $\meanstd{28.4}{6.5}$ & $\meanstd{43.2}{0.8}$ & $\meanstd{17.0}{9.9}$ & 4.3 \\
        C-PCFG* &
        $\meanstd{36.0}{1.2}$ & $\meanstd{38.7}{6.6}$ & $\meanstd{56.3}{2.1}$ & $\meanstd{45.0}{1.1}$ & $\meanstd{43.5}{1.2}$ & $\meanstd{45.2}{0.5}$ & $\mathbf{\meanstd{44.9}{1.5}}$ & $\meanstd{30.5}{4.2}$ & $\meanstd{43.8}{1.3}$ & $\meanstd{33.0}{15.4}$ & 3.2 \\
        TN-PCFG* &
        $\meanstd{36.0}{3.0}$ & $\meanstd{39.2}{5.0}$ & $\meanstd{57.7}{4.2}$ & $\meanstd{39.1}{4.1}$ & $\meanstd{47.1}{1.7}$ & $\meanstd{39.2}{10.7}$ & $\meanstd{43.1}{1.1}$ & $\meanstd{35.4}{2.8}$ & $\mathbf{\meanstd{48.6}{3.1}}$ & $\mathbf{\meanstd{40.0}{4.8}}$ & 3.1 \\
        FGG-TNPCFG$^{\dag}$ & 
        $\meanstd{38.4}{7.3}$ & $\meanstd{31.0}{8.4}$ & $\meanstd{59.6}{7.7}$ & $\meanstd{43.9}{3.1}$ & $\meanstd{48.0}{1.4}$ & $\meanstd{46.2}{4.1}$ & $\meanstd{42.2}{0.7}$ & $\meanstd{31.5}{4.0}$ & $\meanstd{41.6}{4.3}$ & $\mathbf{\meanstd{40.0}{0.6}}$ & 3.0 \\
        Ours &
        $\mathbf{\meanstd{45.9}{0.3}}$ & $\mathbf{\meanstd{46.1}{0.9}}$ & $\mathbf{\meanstd{69.7}{0.9}}$ & $\mathbf{\meanstd{50.5}{0.5}}$ & $\mathbf{\meanstd{49.1}{0.3}}$ & $\mathbf{\meanstd{49.5}{0.2}}$ & $\meanstd{43.7}{0.2}$ & $\mathbf{\meanstd{42.1}{0.3}}$ & $\meanstd{47.9}{0.3}$ & $\meanstd{33.4}{1.2}$ & \textbf{1.4} \\
    \hline
    \end{tabular}
    \end{adjustbox}
    \caption{Performance (S-F1 score) in multilingual parsing on CTB and SPMRL datasets. * indicates reported by~\citet{yang2021pcfgs}, $\dagger$ indicates results reproduced using open-sourced code provided by~\citet{yang2022dynamic}.}
    \label{tab:multilingual-evaluation}
\end{table*}

Table~\ref{tab:multilingual-evaluation} shows the performance of explicit grammar-based UNGIs in multilingual parsing on CTB and SPMRL data. The results indicate that our model significantly outperforms other models across most languages. While a few languages, such as Hungarian, Polish, and Swedish, show second or third-best performance, our model ranks the highest overall.
Note that our method has consistently and significantly lower variance compared to all other methods, which indicate that it is a more reliable estimator. 

\subsection{In-depth Analysis}
\paragraph{Performance Consistency about Randomness}
We investigate the impact of our method on the high variance problem, which is the main limitation of UNGIs.
To observe the variance change caused solely by parse-focusing, we also evaluate the variance using the generated sets of parse trees as the focusing-bias: randomly, left-only branched, and right-only branched.
As shown in Table~\ref{tab:bias-injection}, all of the variances of S-F1 scores are significantly reduced compared to those of FGG-TNPCFG, while their negative log likelihoods exhibit a similar bias.
This means that sentence-wise parse-focusing reduces variance without depending on the selected parses.
The results indicate that applying parse-focusing reduces the variance problem and indirectly supports the idea that the cause is due to SOA under uncorrelated S-F1 and NLL.
Moreover, they show that focusing-bias from pre-trained parsers can lead to higher performance than algorithmically generated parses.

\begin{table}[t!]
    \centering
    \small
    \begin{adjustbox}{max width=\columnwidth}
    \begin{tabular}{lcc}
    \hline
        Bias & S-F1 & NLL \\
    \hline
        LBranching & $\meanstd{8.9}{0.0}$ & $\meanstd{116.5}{2.3}$ \\
        RBranching & $\meanstd{39.7}{0.0}$ & $\meanstd{116.1}{1.6}$ \\
        Random & $\meanstd{34.9}{1.2}$ & $\meanstd{113.5}{1.0}$ \\
        Structformer & $\meanstd{58.9}{0.2}$ & $\meanstd{112.0}{0.9}$ \\
        NBL-PCFG & $\meanstd{64.7}{0.6}$ & $\meanstd{111.5}{0.7}$ \\
        FGG-TNPCFG & $\meanstd{65.3}{0.4}$ & $\meanstd{113.4}{1.3}$ \\
        DIORA & $\meanstd{50.0}{1.4}$ & $\meanstd{111.6}{0.3}$ \\
    \hline
        FGG-TNPCFG$^{\dagger}$ & $\meanstd{57.4}{6.0}$ & $\meanstd{111.9}{1.1}$ \\
    \hline
    \end{tabular}
    \end{adjustbox}
    \caption{S-F1 and NLL for parse-focused grammar by each focusing-bias. LBranching and RBranching refer to left-branching and right-branching, respectively. Random and pre-trained biases are generated once and used for all runs. FGG-TNPCFG$^{\dagger}$ denotes reproduced performance to compare variance.}
    \label{tab:bias-injection}
\end{table}

\paragraph{Performance and Variance by Symbol Size}
\begin{table}[htbp!]
    \small
    \centering
    \begin{tabular}{rrcc}
    \hline
        \multirow{2}{*}{Nonterminal} & \multirow{2}{*}{Preterminal} & \multicolumn{2}{c}{S-F1} \\
        \cline{3-4}
        & & Ours & FGG \\ 
    \hline
        1 & 2 & $\meanstd{44.3}{0.1}$ & $\meanstd{39.7}{0.0}$ \\
        5 & 10 & $\meanstd{60.4}{1.9}$ & $\meanstd{41.6}{9.6}$ \\
        15 & 30 & $\meanstd{65.5}{0.7}$ & $\meanstd{43.2}{0.8}$ \\
        30 & 60 & $\meanstd{67.4}{0.9}$ & $\meanstd{51.2}{3.1}$\\
        250 & 500 & $\meanstd{69.7}{0.9}$ & $\meanstd{54.3}{3.9}$\\
        4500 & 9000 & $\meanstd{69.6}{0.7}$ & $\meanstd{57.4}{6.0}$\\
    \hline
    \end{tabular}
    \caption{The S-F1 score for each number of symbols for FGG-TNPCFGs and induced grammars by parse-focusing.}
    \label{tab:num-of-sym}
\end{table}

In Section~\ref{sec:simplicity}, we show that parse-focusing relaxes the low diversity of unique rules in conventional methods.
Here, we demonstrate how the relaxation from parse-focusing affects the increase in performance according to changes in symbol size.
In Table~\ref{tab:num-of-sym}, although our method boosts performance in the larger model, it significantly enhances performance compared to FGG-TNPCFGs in the smaller grammar model. 
These results imply that our method can stably induce more compact grammars by reducing SSB while achieving more accurate models using the same model capacity. 

\paragraph{Rule Utilization}
\begin{figure}[t!]
    \small
    \centering
    \includegraphics[
        width=\columnwidth
    ]{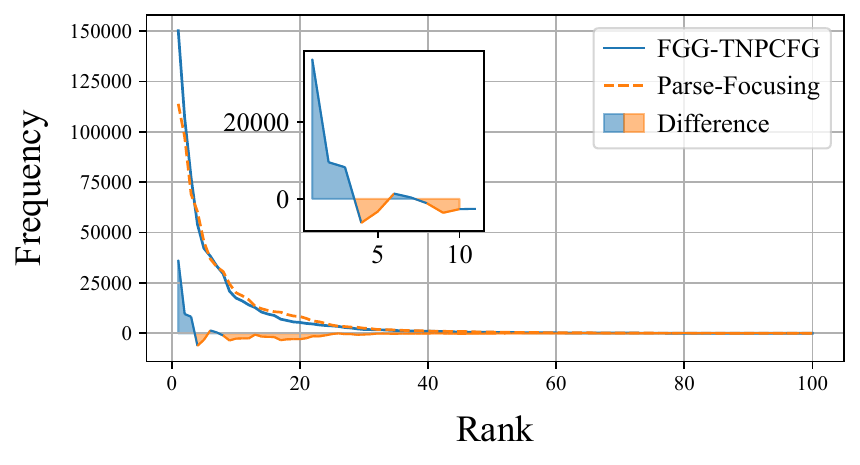}
    \caption{Frequency distribution sorted in descending order for contained rules in the parsed training dataset. The colored fill area represents the frequency difference between FGG-TNPCFGs and our method with 5 nonterminals. The small box zooms in on the top 10 rules for differences.}
    \label{fig:parse-rule-frequency}
\end{figure}

We show that parse-focusing reduces the extreme reliance on a small number of rules in FGG-TNPCFG.
Figure~\ref{fig:parse-rule-frequency} shows the sorted frequency of all rules in a grammar. 
After applying our method, the frequency of the most frequent three rules decreases from 46.1\% to 38.7\%, and the reduced frequency is redistributed to other less frequently used rules.
These results imply the effective enhancement of rule utilization.

\begin{figure}
    \centering
    \includegraphics[
        width=\columnwidth
    ]{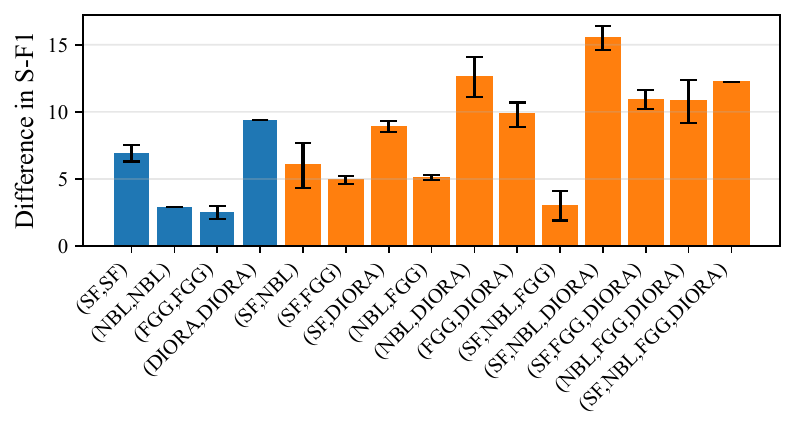}
    \caption{The difference between the mean of S-F1 scores for pre-trained parsers and the parse-focused model trained by those parsers. The blue bars indicate homogeneous parsers, and the orange bars indicate heterogeneous parsers. The raw S-F1 scores are shown in Figure~\ref{fig:homo_hetero_total}.}
    \label{fig:homo_hetero_diff}
\end{figure}

\paragraph{Practical Multi-Parser Combinations}
We show the significance of using heterogeneous multi-parsers that have different model-specific inherent biases.
Figure~\ref{fig:homo_hetero_diff} shows that the performance increase in heterogeneous parsers is generally more considerable compared to homogeneous ones.
These parsers exhibit more significant variances in performance, which supports the idea that the inherent biases of models are offset.
Consequently, the results implies that heterogeneity is a practically helpful strategy.

\section{Related Works}
\paragraph{TN-PCFG/FGGs} 
\citet{yang2021pcfgs} and \citet{yang2022dynamic} verify that overparameterization leads to better grammars with many symbols. 
Conversely, we highlight the inefficiency in learning UGIs caused by simplicity and ambiguity issues. Solving this inefficiency enables learning even a small grammar with significantly high performance and low variance. 

\paragraph{S-DIORA}
\citet{drozdov-etal-2020-unsupervised} shows that structural ambiguity between subtrees can affect locally greedy training, and that this problem can be resolved by single tree encoding through beam search. This work supports the impact of the ambiguity issue, although our research addresses the issue of learning explicit grammars in UGIs differently.

\paragraph{Ensemble Distillation}
\citet{shayegh2023ensemble} introduce a variant of the CYK algorithm that gains refined parse trees by combining trees from the different pre-trained parsers.
Our approach differs in searching for structure through training trees in a relaxed condition using soft weights from the count of spans.
It utilizes the model's search capability.
In addition, it is different that we resolve the optimization problem for the loss function.

\paragraph{Tree-shape Uncertainty}
The tree-shape uncertainty referred to in \citet{ishii-miyao-2023-tree} is the existence of similar grammars that assign different tree structures for a given sentence.
In our work, the issue is analyzed from an optimization perspective, highlighting how undifferentiated optima can cause unstable learning.

\paragraph{What Do Implicit Grammar Models Learn?}
\citet{latent2018williams} investigate how consistently latent tree learning models learn in the presence of randomness.
In our work, we focus on identifying the causes of inconsistency and its negative effects, which are primarily due to likelihood loss over all parses in UGI.

\section{Conclusion}
In this paper, we tackle the challenges of high variance and overly simplistic parse in unsupervised neural grammar induction (UNGI). We identify \textit{structural optimization-ambiguity} and \textit{structural simplicity bias} as primary causes, arising from UGI loss on all possible parses of all sentence parses. We introduce a parse-focusing approach that narrows down the parse set per sentence and incorporates biases from parses generated by pre-trained parsers on the same dataset. This method has been shown to effectively mitigate these issues, significantly surpassing state-of-the-art methods in both English and multilingual parsing benchmarks. Our contributions facilitate the consistent learning of more compact, explicit, accurate, and therefore interpretable grammar models in unsupervised learning environment.

\section*{Acknowledgements}

This work was supported by the National Research Foundation of Korea (NRF) grant funded by the Korea government (MSIT) (No.2022R1A2C2012054, Development of AI for Canonicalized Expression of Trained Hypotheses by Resolving Ambiguity in Various Relation Levels of Representation Learning).

\section*{Limitations}
Our work is limited to offline learning due to additional computational costs for pre-training parsers and for the hyper-parameter search of better combinations. In the future, more light and adaptive structural bias generation is required. 
%


\bibliography{custom}

\appendix

\section{Appendices}

\subsection{Detail Proof of Structural Optimization Ambiguity with the Inside Algorithm}
\label{appendix:ambiguity-proof}

\begingroup
\renewcommand{\arraystretch}{1.2} 
\begin{table*}
    \centering
    $\begin{array}{@{}cL@{}}
        \hline
        \multicolumn{1}{l}{\textbf{Conventional inside algorithm}}
        & \multicolumn{1}{l}{}\\
        \hline
        \begin{aligned}
            I_i(1,l)&=\sum_{i}^{n}\sum_{j}^{n}\sum_{k}^{n}\prob{N_i}{N_jN_k}I_j(1,p)I_k(p+1,l)\\
        \end{aligned}
        & eq:conv-inside-algorithm \\
        \hline
        \multicolumn{1}{l}{\textbf{Inside algorithm in Unsupervised Neural Grammar Induction}} 
        & \multicolumn{1}{l}{}\\
        \hline
        \begin{aligned}
            I_i(1,l)&=\sum_{j}^{n}\sum_{k}^{m}\prob{N_i}{N_jT_k}I_j(1,l-1)\prob{T_k}{w_l}+\prob{N_i}{T_kN_j}\prob{T_k}{w_1}I_j(2,l)\\
            &\quad+\sum_{j}^{n}\sum_{k}^{n}\sum_{a=2}^{n-2}\prob{N_i}{N_jN_k}I_j(1,a)I_k(a+1,l)\\
        \end{aligned}
        & eq:variant-inside-algorithm \\
        \hline
        \multicolumn{1}{l}{\textbf{Inside algorithm in UNGI with one preterminal symbol}} 
        & \multicolumn{1}{l}{}\\
        \hline
        \begin{aligned}
            I_i(1,l)&=\sum_{j}^{n}\prob{N_i}{N_jT}I_j(1,l-1)\prob{T}{w_l}+\prob{N_i}{TN_j}\prob{T}{w_1}I_j(2,l)\\
            &\quad+\sum_{j}^{n}\sum_{k}^{n}\sum_{a=2}^{n-2}\prob{N_i}{N_jN_k}I_j(1,a)I_k(a+1,l)\\
        \end{aligned}
        & eq:inside-one-symbol \\
        \hline
        \multicolumn{1}{l}{\textbf{Inside probability for span width 3}} 
        & \multicolumn{1}{l}{}\\
        \hline
        \begin{aligned}
            I_{G,i}(1,l)&=\sum_{j}^{n}\prob[G]{N_i}{N_jT}I_{G,j}(1,l-1)\prob[G]{T}{w_l}+\prob[G]{N_i}{TN_j}\prob[G]{T}{w_1}I_{G,j}(2,l)\\
        \end{aligned}
        & eq:inside-width-3 \\
        \hline
        \multicolumn{1}{l}{\textbf{Equivalence between two different grammars}} 
        & \multicolumn{1}{l}{}\\
        \hline
        \begin{aligned}
            &\prob[G_1]{N_i}{N_jT}I_{G_1,j}(1,r-1)\prob[G_1]{T}{w_r}+\prob[G_1]{N_i}{TN_j}\prob[G_1]{T}{w_1}I_{G_1,j}(2,r)\\
            &=\prob[G_2]{N_i}{N_jT}I_{G_2,j}(1,r-1)\prob[G_2]{T}{w_r}+\prob[G_2]{N_i}{TN_j}\prob[G_2]{T}{w_1}I_{G_2,j}(2,r)\\
        \end{aligned}
        & eq:equivalent-prob \\
        \hline
        \begin{aligned}
            &\prob[G]{N_i}{N_jT}+\prob[G]{N_i}{TN_j}\alpha\\
            &=\prob[G']{N_i}{N_jT}+\prob[G']{N_i}{TN_j}\alpha
        \end{aligned} \text{}
        & eq:equivalent-prob-2\\
        \hline
    \end{array}$
    \caption{The equation of inside algorithm and derived equation for proof.}
    \label{tab:inside-algorithm}
\end{table*}
\endgroup

We incorporate the three stipulated conditions into the conventional inside algorithm as per Equation~\ref{eq:conv-inside-algorithm}: 1) distinguishing between nonterminals and preterminals, 2) allowing all symbols to occupy identical positions in parse trees, and 3) encompassing all structurally possible binary trees for given sentences. Therefore, we can reformulate the inside algorithm as shown in Equation~\ref{eq:variant-inside-algorithm}.

We examine a special case where a grammar contains only one preterminal symbol, as detailed in Equation~\ref{eq:inside-one-symbol}, to investigate the potential for structural optimization ambiguity. We recursively derive the conditions leading to ambiguity.

Before we begin the proof, let us consider two grammars, $G_1$ and $G_2$. The relationship between them is defined as follows:
\begin{gather*}
\prob[G_1]{N_i}{N_jT} \neq \prob[G_2]{N_i}{N_jT} \\
\prob[G_1]{N_i}{TN_j} \neq \prob[G_2]{N_i}{TN_j} \\
\forall r \in R^{c},\ p_{G_1}(r) = p_{G_2}(r)
\end{gather*}
Here, $N_i,N_j$ represent nonterminal symbols, $T$ denotes a preterminal symbol, $\prob[G_1]{N_i}{N_jT}$ represents the probability of the rule $N_i\rightarrow N_jT$ assigned by $G_1$, and $R^{c}$ is the complement set of rules that do not involve $N_i\rightarrow N_jT$ and $N_i\rightarrow TN_j$.

Initially, we can directly determine the probability of a span with width $w=2$:
\begin{equation}
\resizebox{\columnwidth}{!}{%
    $I_{G,i}(p,p+1)=\prob[G]{N_i}{TT}\prob[G]{T}{w_p}\prob[G]{T}{w_{p+1}}$%
    }
\end{equation}
For both grammars $G_1$ and $G_2$, the equality $I_{G_1,i}(p,p+1)=I_{G_2,i}(p,p+1)$ consistently holds for all $p$ because the probability of a span with width  $w=2$ does not involve the rules $N_i\rightarrow N_jT$ and $N_i\rightarrow TN_j$.

For spans of width $w=3$, we refer to Equation~\ref{eq:inside-width-3}.
We set the inside probabilities for grammars $G_1$ and $G_2$ to be equal, such that $I_{G_1,i}(p,p+2)=I_{G_2,i}(p,p+2)$.
Noting that only the probabilities of $N_i\rightarrow N_jT$ and $N_i\rightarrow TN_j$ differ, we see that the other terms cancel out, as shown in Equation~\ref{eq:equivalent-prob}.
Therefore, Equation~\ref{eq:equivalent-prob} can be simplified to Equation~\ref{eq:equivalent-prob-2}, given that $I_{G_1,i}(p,p+1)=I_{G_2,i}(p,p+1)$. Consequently:
\begin{equation}
    \alpha=\frac{\prob[G_1]{T}{w_1}I_{G_1,j}(2,r)}{I_{G_1,j}(1,r-1)\prob[G_1]{T}{w_r}}
\end{equation}
It must be noted that $\prob[G_1]{T}{w_1}I_{G_1,j}(2,r)=I_{G_1,j}(1,r-1)\prob[G_1]{T}{w_r}$ since the preterminal symbol $T$ is unified, leading to $\alpha=1$.

This process can be applied recursively to spans of any width. As a result, $\prob[G_1]{N_i}{N_jT}+\prob[G_1]{N_i}{TN_j}=\prob[G_2]{N_i}{N_jT}+\prob[G_2]{N_i}{TN_j}$ if and only if $I_{G_1,i}(p,q)=I_{G_2,i}(p,q)$.  In simpler terms, the sentence probabilities calculated by $G_1$ and $G_2$ will be identical as long as the sum of probabilities $\prob[G_1]{N_i}{N_jT}+\prob[G_1]{N_i}{TN_j}$ matches that of $G_2$. 

\subsection{Structural Simplicity Bias in Multilingual}
\label{appendix:rule_util_multilingual}

\begin{figure*}
    \centering
    \includegraphics[
        width=\textwidth
    ]{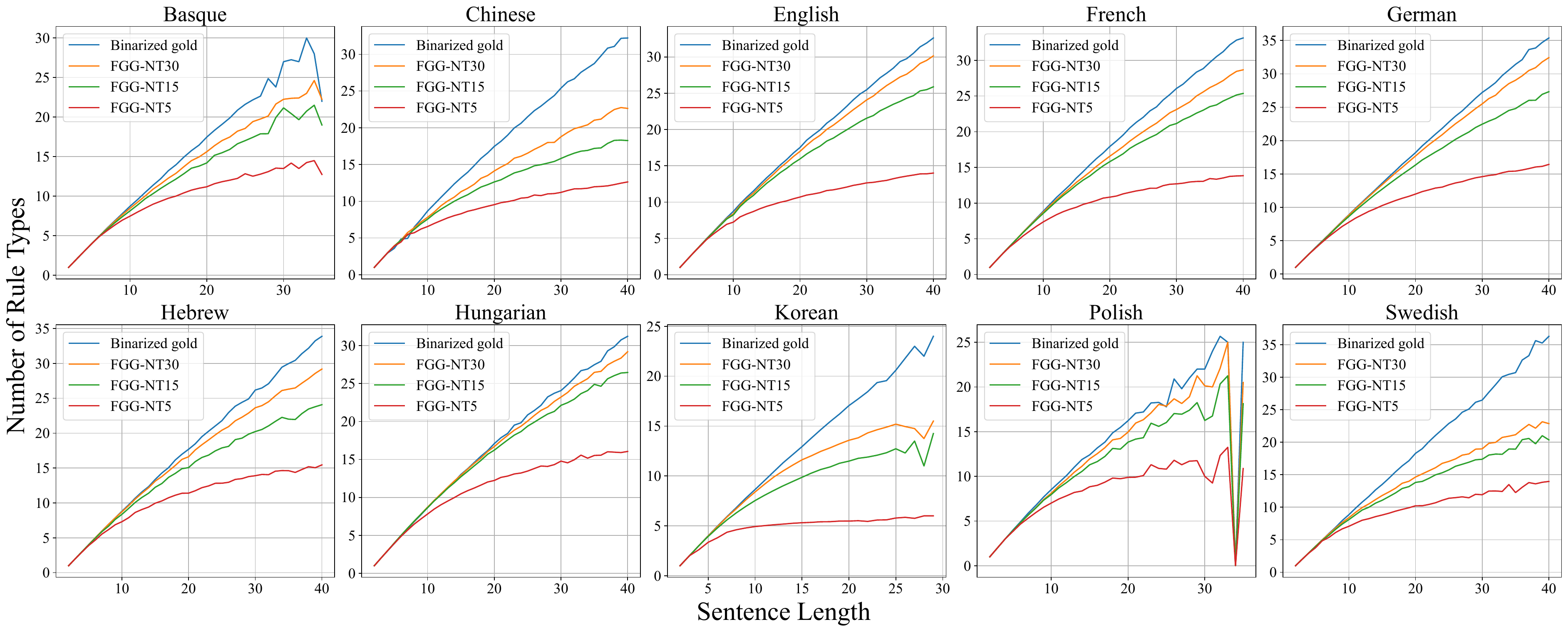}
    \caption{Average of the number of unique rules in each tree for CTB and SPMRL datasets. Binarized gold is the }
    \label{fig:rule_util_multi}
\end{figure*}

We demonstrate that structural simplicity bias is observed in all language datasets.
For this, we train and evaluate FGG-TNPCFGs with 5, 15, and 30 nonterminals using CTB and SPMRL datasets.
In Figure~\ref{fig:rule_util_multi}, it is shown that in all languages, the average number of unique rules decreases as the number of symbols decreases, and verifying that structural simplicity bias is a general issue occurring in all languages, not just limited to PTB.

\subsection{Qualitative analysis of tree structure}
\label{appendix:qa_simplicity}

\begin{figure*}
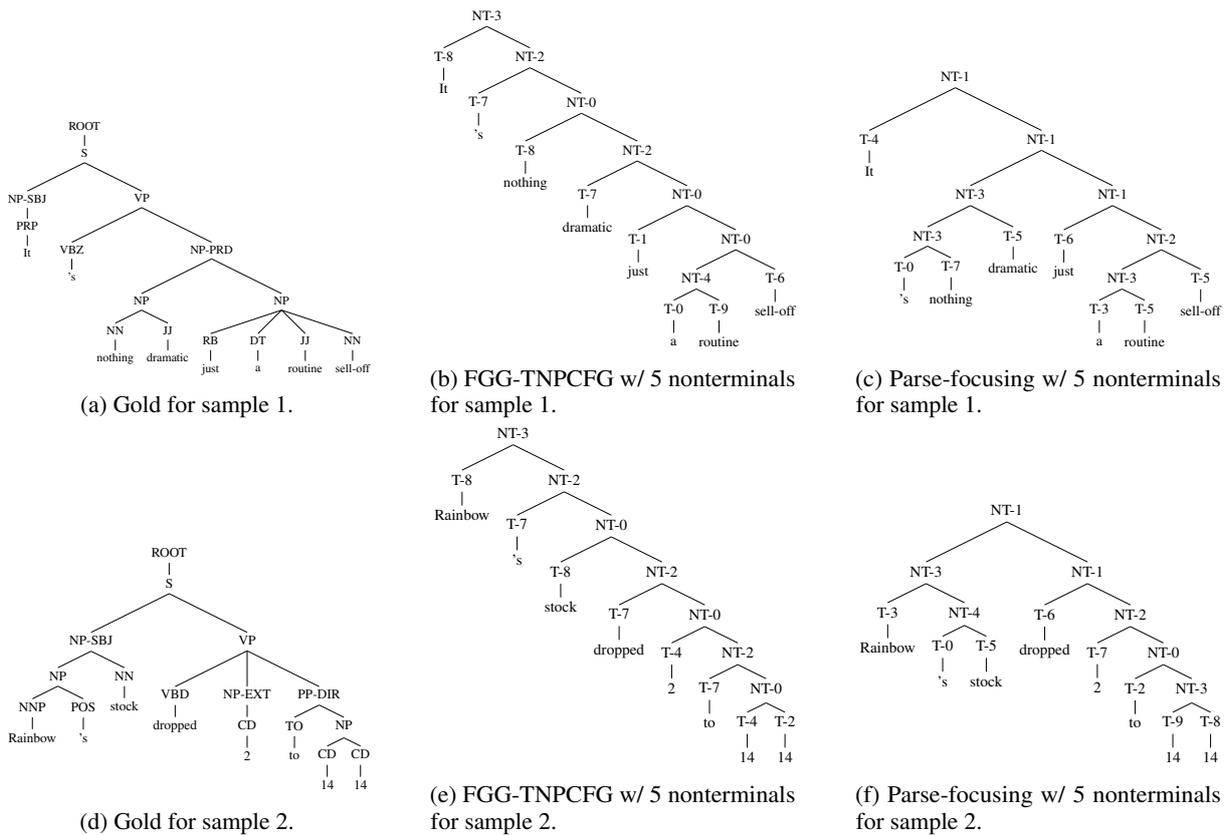

\centering
    \begin{subfigure}[b]{0.3\textwidth}
        \resizebox{\textwidth}{!}{
        \Tree [.ROOT
        [.S
          [.NP-SBJ [.PRP It ] ]
          [.VP
            [.VBZ 's ]
            [.NP-PRD
              [.NP [.NN nothing ] [.JJ dramatic ] ]
              [.NP
                [.RB just ]
                [.DT a ]
                [.JJ routine ]
                [.NN sell-off ] ] ] ] ] ]
        }
        \caption{Gold for sample 1.}
    \end{subfigure}
    \hfill
    \begin{subfigure}[b]{0.3\textwidth}
        \resizebox{\textwidth}{!}{
        \Tree [.NT-3
        [.T-8 It ]
        [.NT-2
          [.T-7 's ]
          [.NT-0
            [.T-8 nothing ]
            [.NT-2
              [.T-7 dramatic ]
              [.NT-0
                [.T-1 just ]
                [.NT-0
                  [.NT-4 [.T-0 a ] [.T-9 routine ] ]
                  [.T-6 sell-off ] ] ] ] ] ] ]
        }
        \caption{FGG-TNPCFG w/ 5 nonterminals for sample 1.}
    \end{subfigure}
    \hfill
    \begin{subfigure}[b]{0.3\textwidth}
        \resizebox{\textwidth}{!}{
        \Tree [.NT-1
        [.T-4 It ]
        [.NT-1
          [.NT-3
            [.NT-3 [.T-0 's ] [.T-7 nothing ] ]
            [.T-5 dramatic ] ]
          [.NT-1
            [.T-6 just ]
            [.NT-2
              [.NT-3 [.T-3 a ] [.T-5 routine ] ]
              [.T-5 sell-off ] ] ] ] ]
        }
        \caption{Parse-focusing w/ 5 nonterminals for sample 1.}
    \end{subfigure}
    \bigskip
    \begin{subfigure}[b]{0.3\textwidth}
        \resizebox{\textwidth}{!}{
        \Tree [.ROOT
        [.S
          [.NP-SBJ [.NP [.NNP Rainbow ] [.POS 's ] ] [.NN stock ] ]
          [.VP
            [.VBD dropped ]
            [.NP-EXT [.CD 2 ] ]
            [.PP-DIR [.TO to ] [.NP [.CD 14 ] [.CD 1\/4 ] ] ] ] ] ]
        }
        \caption{Gold for sample 2.}
    \end{subfigure}
    \hfill
    \begin{subfigure}[b]{0.3\textwidth}
        \resizebox{\textwidth}{!}{
        \Tree [.NT-3
        [.T-8 Rainbow ]
        [.NT-2
          [.T-7 's ]
          [.NT-0
            [.T-8 stock ]
            [.NT-2
              [.T-7 dropped ]
              [.NT-0
                [.T-4 2 ]
                [.NT-2 [.T-7 to ] [.NT-0 [.T-4 14 ] [.T-2 1\/4 ] ] ] ] ] ] ] ]
        }
        \caption{FGG-TNPCFG w/ 5 nonterminals for sample 2.}
    \end{subfigure}
    \hfill
    \begin{subfigure}[b]{0.3\textwidth}
        \resizebox{\textwidth}{!}{
        \Tree [.NT-1
        [.NT-3 [.T-3 Rainbow ] [.NT-4 [.T-0 's ] [.T-5 stock ] ] ]
        [.NT-1
          [.T-6 dropped ]
          [.NT-2
            [.T-7 2 ]
            [.NT-0 [.T-2 to ] [.NT-3 [.T-9 14 ] [.T-8 1\/4 ] ] ] ] ] ]
        }
        \caption{Parse-focusing w/ 5 nonterminals for sample 2.}
    \end{subfigure}
\caption{Induced parse trees for two sentences, "It’s nothing dramatic just a routine sell-off." and "Rainbow’s stock dropped 2 to 14 1\/4."}
\label{fig:qa_simplicity}
\end{figure*}

In Figure~\ref{fig:qa_simplicity}, we qualitatively analyze the structural differences between parses induced by different grammars (Gold, FGG-TNPCFGs and parse-focused N-PCFG) with 5 nonterminal symbols for the same sentence.
We verify that the FGG-TNPCFGs utilize a small number of rule types to induce parses, which leads a different structure compared to the gold parse.

\subsection{Algorithm of Proposed Method}
\label{appendix:alg-method}
Algorithm~\ref{alg:overview} presents the pseudocode for our entire methodology. We calculate the weights to be applied in the Inside Algorithm using softmax, based on the frequency of each span within a sentence as provided by the parse trees from pre-trained parsers.

\begin{algorithm}
\small
\caption{Parse-Focusing Inside algorithm}\label{alg:overview}
\begin{algorithmic}[1]
\Function{Parse-Focusing}{$s\in S$, $\mathcal{T}_{p}$}
\State $\mathcal{T}_{p}(s) \gets \mathcal{T}_p$ \Comment{Subset of parse trees}
\State $C(s) \gets$ \Call{CountSpans}{$\tau_{p_i}$, $|s|$}
\State $W(s) \gets \cfrac{\exp{f(c(p,q)|\tau_{G_p}(s_i))}}{\sum_{p,q}\exp f(c(p,q)|\tau_{G_p}(s_i))}$
\For{$w\gets 2$ to $|s|$}
  \For{$n\gets 0$ to $|s|$}
    \State $I_i(n,n+w) \gets \newline 
    \sum_{j,k}\sum_{m}p(i,j,k)I_j(n,m)I_k(m+1,n+w)$ 
    \State $I_i(n,n+w) \gets w_s(n,n+w)I_i(n,n+w)$
  \EndFor
\EndFor
\State \Return $p(s)$
\EndFunction
\Function{CountSpans}{$\tau_{p_i}$, $|s|$}
  \State Initialize $c(i,j)=0$, for all $0\leq i,j\leq |s|$
  \For{$i \gets 0$ to $|s|$}
    \For{$j \gets i+2$ to $|s|$}
      \ForEach{$\tau_{p_i}\in \Tau_{p}$}
        \If{span $(i,j)$ in $\tau_{p_i}$}
          \State $c(i,j) \gets c(i,j)+1$
        \EndIf
      \EndFor
    \EndFor
  \EndFor
  \State \Return $\mathcal{C}(s)$
\EndFunction
\end{algorithmic}
\end{algorithm}

\subsection{Quality of Focusing-Bias by the number of Multiple Parsers}
\begin{figure}[ht!]
    \small
    \centering
    \includegraphics[
        width=\columnwidth
    ]{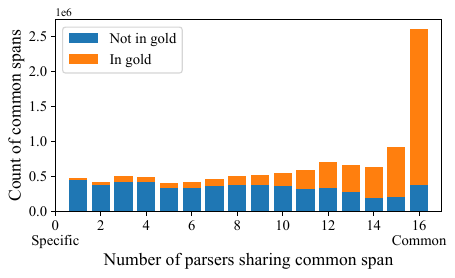}
    \caption{Frequency ($10^6$ scale) of common parse spans in gold parse trees by the number of combined parsers. Using more parsers shows significantly more spans in gold parse trees without supervision.}
    \label{fig:common-part-precision}
\end{figure}

In Figure~\ref{fig:common-part-precision}, the common structures across the selected parsers are more similar to gold parse trees.
This gives the intuition that assigning strong weights to parts that commonly appear across multiple parsers will lead to better performance.

\subsection{Experiment Details}
\label{appendix:exp_details}

\paragraph{Dataset Detail}
We follow the standard setup and preprocessing for the PTB\footnote{The license of PTB is LDC User Agreement for Non-Members. https://catalog.ldc.upenn.edu/LDC99T42}, utilizing sections 02-21 as the training set, section 22 as the validation set, and section 23 as the test set excluding punctuation and trivial constituents. The most frequent 10,000 words were selected as the vocabulary, and the remaining words were treated as \verb|<unk>|. To process the data, we followed the pipeline used by models employing base models and parsers as outlined in \cite{kim2019compound,yang2022dynamic,shen2021structformer,yang2021neural,drozdov2019unsupervised}.
We also use CTB\footnote{The license of CTB is LDC User Agreement for Non-Members. https://catalog.ldc.upenn.edu/LDC2005T01.} and Basque, French, German, Hebrew, Hungarian, Korean, Polish, Swedish dataset in SPMRL\footnote{The license of SPMRL is Creative Commons Attribution 4.0 International License. https://www.spmrl.org/spmrl2013-sharedtask.html}.

\paragraph{Implementation Detail}
To implement our methodology on top of the base model FGG-TNPCFGs, we utilized PyTorch version 2.2~\cite{paszke2019pytorch}. For smooth processing and analysis of tree structures, we employed NLTK\footnote{https://www.nltk.org/}.

\paragraph{Training Detail}
The hyperparameter details are identical to those found in \citet{yang2022dynamic}. The ratio of nonterminal to preterminal symbols is primarily set at 1:2, and to analyze performance based on the number of symbols, experiments were conducted using a minimum of 1 to a maximum of 4500 nonterminal symbols. Training was carried out for up to 10 epochs, with early stopping applied based on validation likelihood. For optimization, the Adam optimizer was utilized with a learning rate of $2e^{-3}$, $\beta_1=0.75$, and $\beta_2=0.999$. The model's hyperparameters were set to a rank size of 1000, a symbol embedding size of 256, and a word embedding size of 200. The same hyperparameters were used for all languages without any specific tuning. All experiments were performed using an NVIDIA RTX 2080ti, and 32 experiments were conducted to verify variance for both the original FGG-TNPCFGs and our method, as shown in Table~\ref{tab:final-performance}, while the rest of the experiments were carried out four times each. Each run takes approximately 3 hours.




\subsection{Focusing-Bias of Single Parser}
\label{appendix:focusing-biase-single-parser}

To validate the learning of various types of focusing-bias in the N-PCFG, we assessed the Intersection over Union (IoU) scores between the pre-trained single parsers and our parse-focused N-PCFGs, as shown in Table~\ref{tab:single_homo_us_parsers}. The results in Table~\ref{tab:single_homo_us_parsers} indicate that our parse-focusing N-PCFGs can replicate the tree structures generated by the pre-trained parsers. They show the highest congruence with FGG-TNPCFGs and the least congruence with Structformer.
Despite this, the Structformer, which has the lowest IoU score, still surpasses the highest IoU score listed in Table~\ref{tab:heteroparsers-iou}. This indirectly indicates the extent of similarity between the parse trees derived from the pre-trained parsers and those obtained from the parse-focused N-PCFG.


\begin{table}[ht!]
    \centering
    \begin{adjustbox}{max width=\columnwidth}
    \begin{tabular}{lccc}
    \hline
        \multirow{2}{*}{Bias} & \multicolumn{2}{c}{S-F1} & \multirow{2}{*}{IoU}\\
        & Pre-trained & Injected & \\
    \hline
        SF & $52.32_{\pm2.25}$ & $56.35_{\pm2.03}$ & $57.18_{\pm2.81}$ \\
        NBL & $52.11_{\pm15.32}$ & $53.04_{\pm16.03}$ & $61.37_{\pm2.39}$ \\
        FGGs & $59.65_{\pm7.73}$ & $58.67_{\pm7.15}$ & $74.49_{\pm6.67}$ \\
    \hline
    \end{tabular}
    \end{adjustbox}
    \caption{S-F1 score of Pre-trained parsers, Parse-focused N-PCFG, and IoU score between their results.}
    \label{tab:single_homo_us_parsers}
\end{table}

\begin{figure}
    \centering
    \includegraphics[
        width=\columnwidth
    ]{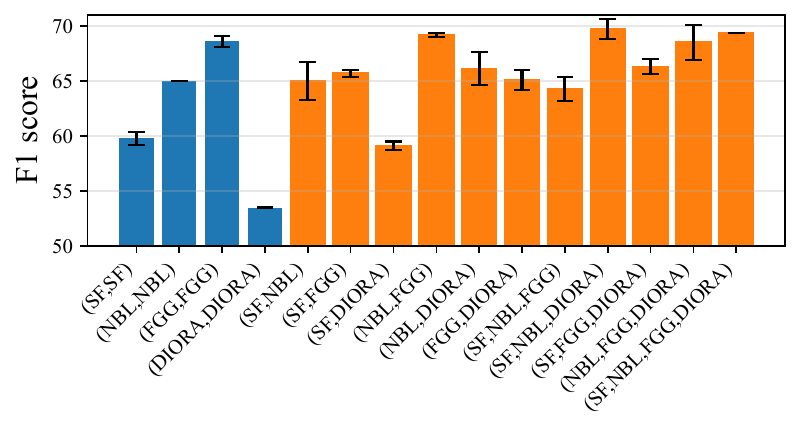}
    \caption{S-F1 scores of the combination of homogeneous and heterogeneous parsers.}
    \label{fig:homo_hetero_total}
\end{figure}

\subsection{Numerical Result for Figure~\ref{fig:homo_hetero_total}}
\label{appendix:num_res_homo_hetero}

In Table~\ref{tab:heteroparsers-total}, the S-F1 and IoU scores for combinations of homogeneous parsers and heterogeneous parsers.
However, heterogeneous parsers show significantly lower IoU scores across all combinations compared to homogeneous parsers.
A lower IoU score signifies that the two grammars make different decisions for more spans, allowing the model to handle a wider variety of structural information in parse-focusing. The model can select the most advantageous structural information from the given options, opening up opportunities for exploration, potentially leading to higher performance.

\begin{table}[t!]
    \small
    \centering
    \begin{tabular}{lcc}
    \hline
        Parser Pair & S-F1 & IoU \\
    \hline
        \multicolumn{3}{l}{Homogeneous (same method, different seeds)} \\
    \hline
        (SF,SF) & $\meanstd{59.8}{0.6}$  & 50.6 \\
        (NBL,NBL) & $\meanstd{65.0}{0.2}$ & 57.1 \\
        (FGG,FGG) & $\meanstd{68.6}{0.5}$ & 40.8 \\
        (DIORA,DIORA) & $\meanstd{53.5}{0.7}$ & 43.9 \\
    \hline
        \multicolumn{3}{l}{Heterogeneous (different method, different seeds)} \\
    \hline
        (SF,NBL) & $\meanstd{65.0}{1.7}$ & 40.3 \\
        (SF,FGG) & $\meanstd{65.7}{0.3}$ & 39.6 \\
        (SF,DIORA) & $\meanstd{59.1}{0.4}$ & 36.2 \\
        (NBL,FGG) & $\meanstd{69.2}{0.2}$ & 40.6 \\
        (NBL,DIORA) & $\meanstd{66.1}{1.5}$ & 32.9 \\
        (FGG,DIORA) & $\meanstd{65.1}{0.9}$ & 34.3 \\
        (SF,NBL,DIORA) & $\meanstd{64.3}{1.1}$ & 21.0 \\
        (SF,FGG,NBL) & $\meanstd{69.7}{0.9}$ & 24.7 \\
        (SF,FGG,DIORA) & $\meanstd{66.3}{0.7}$ & 21.7 \\
        (FGG,NBL,DIORA) & $\meanstd{68.5}{1.6}$ & 20.7 \\
    \hline
    \end{tabular}
    \caption{IoU scores of homogeneous and heterogeneous parser combinations. S-F1 is the test performance of ours with the parser combination on PTB. FGG denotes FGG-TNPCFG.}
    \label{tab:heteroparsers-total}
\end{table}

\subsection{Recent Studies}
\label{appendix:background}

With the dramatic growth of neural networks, various approaches have been developed to integrate the concepts of neural networks and grammars.~\cite{shen2018neural,shen2019ordered} utilized tree structures within their models, hierarchically learning implicit structural information from sentences using tree structures.
Meanwhile, \citet{wang2019tree,shen2021structformer} aimed to leverage structural information within transformers by inferring structural information from sentences and using it to mask the attention matrix to control relationships between words. Notably, \citet{shen2021structformer} proposed a method for simultaneously learning both constituency and dependency parsing.

There have been studies applying structural concepts, either by mimicking grammars using recurrent neural networks as in \citet{dyer-etal-2016-recurrent,kim-etal-2019-unsupervised} or by combining the inside algorithm with the concept of autoencoders as in~\citet{drozdov2019unsupervised}.

In particular, \citet{kim2019compound} proposed neural parameterization, assigning rule probabilities using neural networks to grammar component embeddings. \citet{kim2019compound} utilized explicit grammatical structures and the inside algorithm to induce grammars, thus enabling traditional grammatical approaches. This led to the emergence of lexicalized N-PCFGs ~\cite{zhu2020return,yang2021neural}, which leverage lexicons. However, the inside algorithm still demanded high computational complexity, which limited the expansion of grammar sizes.

\citet{yang2021pcfgs,yang2022dynamic} addressed the high computational complexity of the inside algorithm by utilizing tensor decomposition and FGGs, which allowed for the learning of larger grammars. This approach, which solves issues through the number of parameters, aligns with the methods of \citet{petrov2006learning}.

\end{document}